\documentclass{article}

\usepackage{arxiv}
\usepackage{amsmath}

\usepackage[utf8]{inputenc} 
\usepackage[T1]{fontenc}    
\usepackage[bottom]{footmisc}
\usepackage[hyperfootnotes=false]{hyperref}       
\usepackage{url}            
\usepackage{booktabs}       
\usepackage{amsfonts}       
\usepackage{nicefrac}       
\usepackage{microtype}      
\usepackage{cleveref}       
\usepackage{lipsum}         
\usepackage{graphicx}
\usepackage{natbib}
\usepackage{doi}
\usepackage{float}
\usepackage{comment}

\title{Enhancing stop location detection for incomplete urban mobility datasets}

\newif\ifuniqueAffiliation

\ifuniqueAffiliation 
\author{ 
\href{https://orcid.org/0000-0003-4372-9734}{
        \includegraphics[scale=0.06]{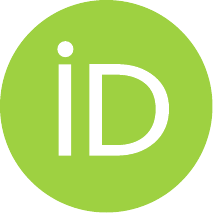}\hspace{1mm}Margherita Bertè}
        \\
        IMT School for Advanced Studies Lucca\\
	Italy\\
	\texttt{margherita.berte@imtlucca.it} \\
	\And
	\href{https://orcid.org/0000-0000-0000-0000}{
        \includegraphics[scale=0.06]{orcid.pdf}\hspace{1mm}Rashid Ibrahimli}\thanks{Department of Data Science, Caja Blanca Datos SL, 28013 Madrid, Spain}
        \\
	Grupo Interdisciplinar de Sistemas Complejos  (GISC)\\
	Department of Mathematics\\
	Carlos III University of Madrid\\
	28911 Leganés, Spain\\
	\texttt{rashid.ibrahimli@whiteboxml.com} \\
    \And
        \includegraphics[scale=0.06]{orcid.pdf}\hspace{1mm}Lars Koopmans \\
	Department of Theoretical Ecology\\
	University of Amsterdam\\
	Amsterdam, The Netherlands \\
	\texttt{l.koopmans@uva.nl} \\
    \And
        \includegraphics[scale=0.06]{orcid.pdf}\hspace{1mm}Pablo Valgañón \\
	Department of Electrical Engineering\\
	Mount-Sheikh University\\
	Santa Narimana, Levand \\
	\texttt{aaaaaa} \\
 \And
        \includegraphics[scale=0.06]{orcid.pdf}\hspace{1mm}Nicola Zomer \\
	Department of Physics and Astronomy ``Galileo Galilei''\\
	University of Padua\\
	Via F. Marzolo 8, 35131 Padua, Italy \\
	\texttt{nicola.zomer@studenti.unipd.it} \\
 \And
        \includegraphics[scale=0.06]{orcid.pdf}\hspace{1mm}Davide Colombi \\
	Cuebiq Inc., Milan, Italy \\
	\texttt{dcolombi@cuebiq.com} \\
}
\else
\usepackage{authblk}

\newbox{\orcid}\sbox{\orcid}{\includegraphics[scale=0.06]{orcid.pdf}} 
\author[1]{%
	\href{https://orcid.org/0000-0003-4372-9734}{\usebox{\orcid}\hspace{1mm}Margherita Bertè
}}
\author[2, 3]{%
	\href{https://orcid.org/0009-0005-4845-1710}{\usebox{\orcid}\hspace{1mm}Rashid Ibrahimli
}}
\author[4]{%
    \usebox{\orcid}\hspace{1mm}Lars Koopmans
}
\author[5]{%
	\usebox{\orcid}\hspace{1mm}Pablo Valgañón
}
\author[6]{%
	\usebox{\orcid}\hspace{1mm}Nicola Zomer
}
\author[7]{%
    {\usebox{\orcid}\hspace{1mm}Davide Colombi}%
}

\affil[1]{IMT School for Advanced Studies Lucca
Lucca, Italy}
\affil[2]{
Department of Mathematics, Carlos III University of Madrid, 28911 Leganés, Spain}
\affil[3]{Department of Data Science, Caja Blanca Datos SL, 28013 Madrid, Spain}
\affil[4]{Department of Theoretical Ecology, University of Amsterdam, Amsterdam, The Netherlands
}
\affil[5]{Department of Condensed Matter Physics, University of Zaragoza, 50009 Zaragoza, Spain
}
\affil[6]{Department of Physics and Astronomy “Galileo Galilei”,
University of Padua, 35131 Padua, Italy
}
\affil[7]{Cuebiq Inc., Milan, Italy
}

\fi

\hypersetup{
pdftitle={Enhancing stop location detection for incomplete urban mobility datasets},
pdfauthor={Margherita Bertè, Rashid Ibrahimli, Lars Koopmans, Pablo Valgañón, Nicola Zomer, Davide Colombi},
pdfkeywords={First keyword, Second keyword, More},
}

\begin{document}
\maketitle

\begin{abstract}
Stop location detection, within human mobility studies, has an impacts in multiple fields including urban planning, transport network design, epidemiological modeling, and socio-economic segregation analysis. However, it remains a challenging task because classical density clustering algorithms often struggle with noisy or incomplete GPS datasets. This study investigates the application of classification algorithms to enhance density-based methods for stop identification. Our approach incorporates multiple features, including individual routine behavior across various time scales and local characteristics of individual GPS points.
The dataset comprises privacy-preserving and anonymized GPS points previously labeled as stops by a sequence-oriented, density-dependent algorithm. We simulated data gaps by removing point density from select stops to assess performance under sparse data conditions. The model classifies individual GPS points within trajectories as potential stops or non-stops.
Given the highly imbalanced nature of the dataset, we prioritized recall over precision in performance evaluation. Results indicate that this method detects most stops, even in the presence of spatio-temporal gaps and that points classified as false positives often correspond to recurring locations for devices, typically near previous stops.
While this research contributes to mobility analysis techniques, significant challenges persist. The lack of ground truth data limits definitive conclusions about the algorithm's accuracy. Further research is needed to validate the method across diverse datasets and to incorporate collective behavior inputs.
\end{abstract}



\section{Introduction}

The advent of mobile devices and positioning technologies in recent decades enables the inference of trajectories from sources like mobile phone logs, social network data, and GPS records. Researchers now can leverage on large-scale datasets from electronics, media, telecommunications companies, and experimental studies to analyze the movements of millions of individuals over extensive periods with high spatial and temporal resolution.

Stop location detection, as part of the exploration of human mobility, allows for the analysis of personal habits and patterns at the community level to address real-world challenges related to urban planning studies~\citep{DeNadai2016}, efficient transport network design~\citep{Ferretti2018} and sustainable cities~\citep{Caitlin2015}. It has also contributed to the understanding of the spread of diseases~\citep{Moritz2020, Oliver2020, Aguilar-Sanchez2022, klein2023forecasting} and socio-economic segregation~\citep{Pappalardo2015, Centellegher2024}. 
Despite its importance, this area has received relatively less attention than research on trajectory reconstruction~\citep{Ramaswamy2004, Aslak2020}. The challenge is further complicated by the inherent noise in mobility data, which is a major obstacle for accurate stop detection methods.

The most commonly used methods for detecting stops in mobility data are based on density principles, identifying stops in areas with high concentrations of GPS points. For instance, density-based clustering algorithms, such as DBSCAN (Density-Based Spatial Clustering of Applications with Noise)~\citep{ester1996density} and OPTICS (Ordering Points To Identify the Clustering Structure)~\citep{ankerst1999optics}, uses concepts like density reachability to identify clusters.

Several other clustering-based methods have been developed specifically for stop extraction, incorporating average GPS point speed~\citep{palma2008clustering} and using trajectory segmentation approaches~\citep{buchin2011segmenting, soares2015grasp} .
Furthermore, those methods have limitations in handling noisy data, identifying isolated stop points with large time intervals, and merging noise-interrupted sequences into stops. 
One of the most widely adopted methods, known as Project Lachesis~\citep{Ramaswamy2004}, initially filters out non-stationary positions and subsequently clusters nearby points within a specified radius and time interval using DBSCAN.
More recently,~\citep{Aslak2020} introduced Infostop, building upon similar filtering techniques to employ the network community detection algorithm Infomap to identify destination points for each individual, even for multi-user traces.

\begin{figure}[ht]
    \includegraphics[width=0.65\textwidth]{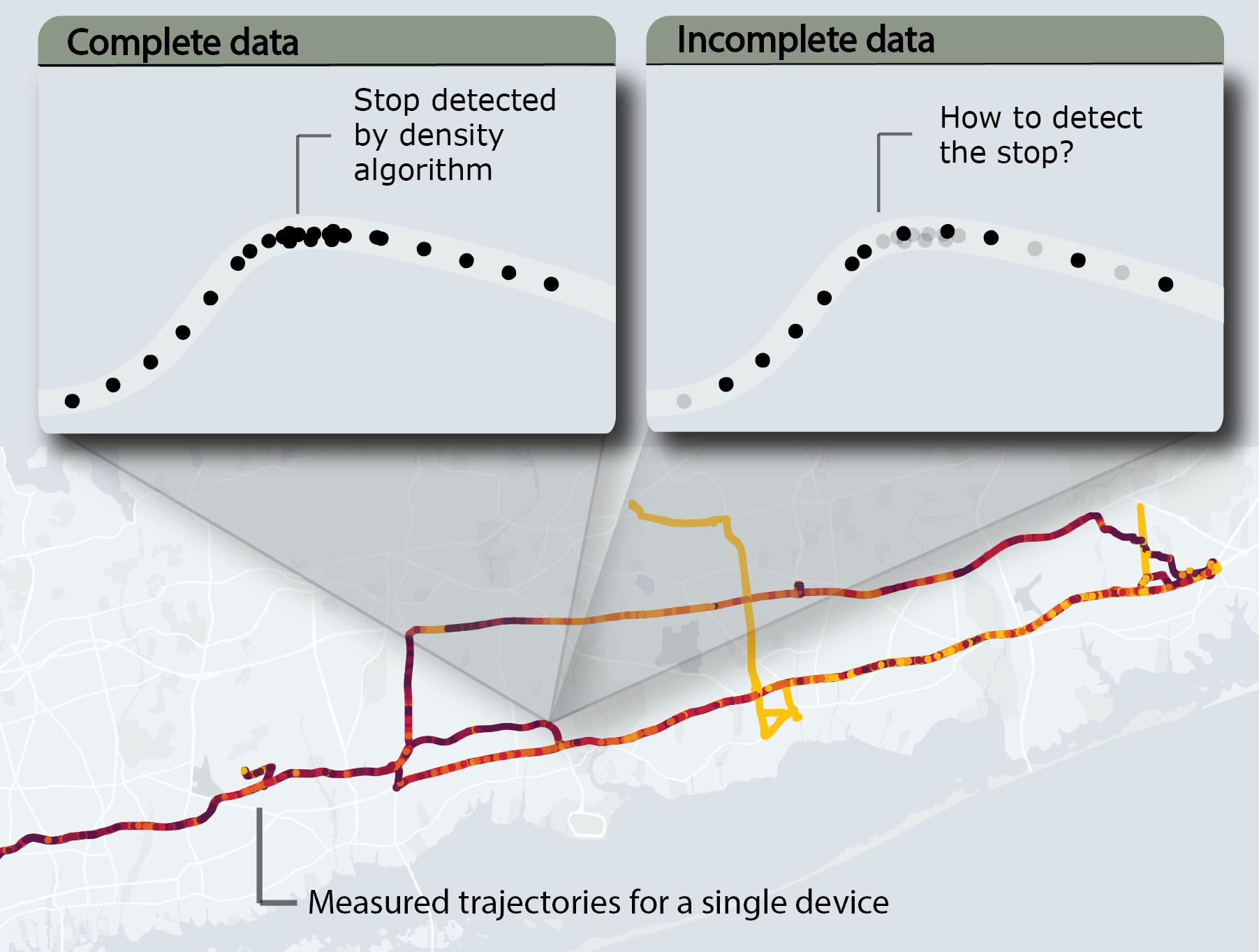}
	\centering
	\caption{This sample trajectory, derived from GPS data, illustrates a common challenge: density-based algorithms struggle to accurately detect stop locations when confronted with noisy or incomplete data, particularly when temporal and spatial gaps are present. (Note: the above figure represents a sample trajectory derived from synthetic points in order to preserve privacy) 
 }
	\label{fig:fig_problem_setup}
\end{figure}

However, using only density related methods may pose risks due to the inherent noise or missing data in GPS records, which can suffer from spatial and temporal gaps. Figure \ref{fig:fig_problem_setup} provides a visual representation of the problem addressed in this study.

Our proposed method addresses this challenge by using classification data from density-based methods to develop a stop-detection model. The model is trained on a feature set primarily derived from individual routines and local characteristics of discrete GPS points. Furthermore, incorporating collective behavioral patterns might bridge gaps created by missing individual data and could offer insights for identifying stop locations beyond an individual's routine~\citep{Bontorin2024a}.




\section{Methods}
\label{sec:methods}

\subsection{Data description}
The data is provided by Cuebiq Inc. through their Data for Good initiative\footnote{\url{https://www.cuebiq.com/about/data-for-good/}}.
The dataset comprises privacy-enhanced (through up-leveling and aggregation) GPS locations, covering the period from February 2024 to March 2024 in the Core-based statistical area (CBSA) of New York-Newark-Jersey City, NY-NJ-PA MSA. Following the approach adopted by Klein et al.~\citep{Klein2023} to ensure data quality, we include only users active from February 1, 2024, participating for at least 20 days per month between February and March 2024, and generating at least 200 location ping per day on average. This filtering results in 3510541 points being included in the dataset. This dataset includes only users who voluntarily opted to share their data anonymously through a GDPR and CCPA compliant process. To enhance and maintain user privacy, sensitive locations such as home and work are obfuscated to the Census Block Group level,. and Sensitive Points of Interest are removed from the dataset entirely The data collection process utilizes the Cuebiq Software Development Kit (SDK), which gathers user locations via GPS and Wi-Fi signals from Android and iOS devices.\\
For the geographic aggregation of points, we use geohashes, a hierarchical spatial data structure that encodes locations using alphanumeric strings. This system subdivides geographic space into grid-shaped buckets, with the location's precision increasing proportionally to the length of the geohash string. This method allows for efficient spatial indexing and aggregation of our dataset while maintaining adjustable levels of geographic precision.
Our data is collected at various geohash levels, with the highest level being up to level 9, depending on the device.
For our analysis, we aggregate the data to hash level 8: an 8-character geohash is approximately 25m$\times$5m in size, which may vary based on the region on the globe (e.g., 38m$\times$19m at the equator)~\citep{geohash}.

Each point is characterized by a unique ID associated with the device, geographical coordinates and geohash of its position, and a timestamp.
To evaluate the robustness of our stop detection model in the presence of data gaps, we implement a synthetic data reduction process on top of a sequence-oriented and density-based stop detection algorithm. This approach allows us to simulate real-world scenarios where data could be incomplete or missing.

\begin{figure}[b!]
    \includegraphics[width=0.6\textwidth]{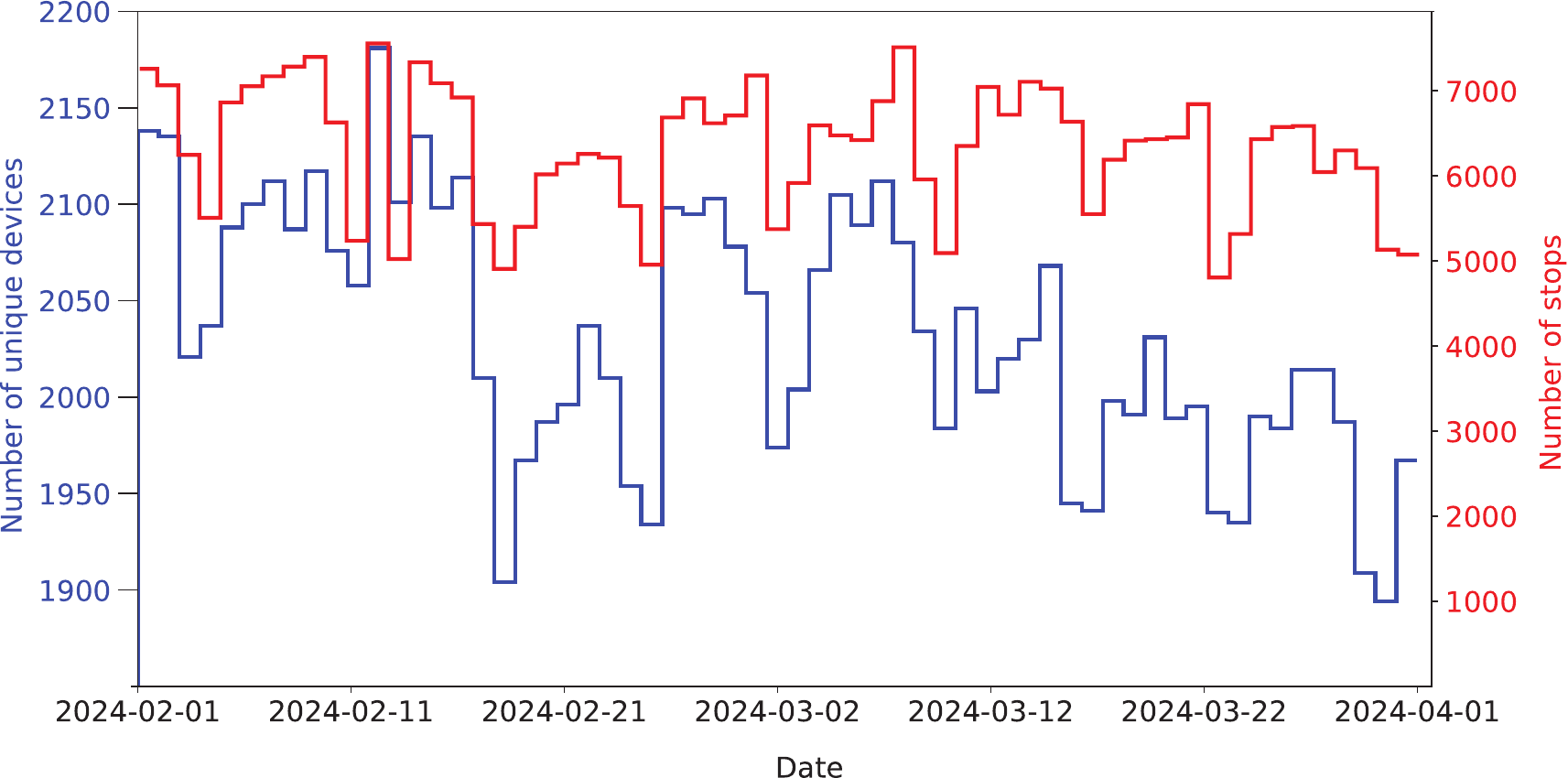}
	\centering
	\caption{Daily trends showing the number of unique devices (blue line) and the number of stops (red line). 
 }
	\label{fig:fig_trend}
\end{figure}

Our dataset comprises individual location pings, each annotated with labels indicating whether the sequence-oriented and density-dependent algorithm identified them as part of a stop. To simulate temporal and spatial gaps in the data, we systematically remove the entire point cloud for 10\% of the detected stops, retaining only two points from each stop. This data reduction process was conducted considering several factors, including device persistence, geohash type, stop classification given by the device traffic in that area, and day of the week.
This synthetic data manipulation allows us to assess the performance of our model under conditions of data sparsity and provides a controlled environment to compare our method against existing approaches that may be less robust to missing data.

\begin{figure}[t!]
    \includegraphics[width=0.52\textwidth]{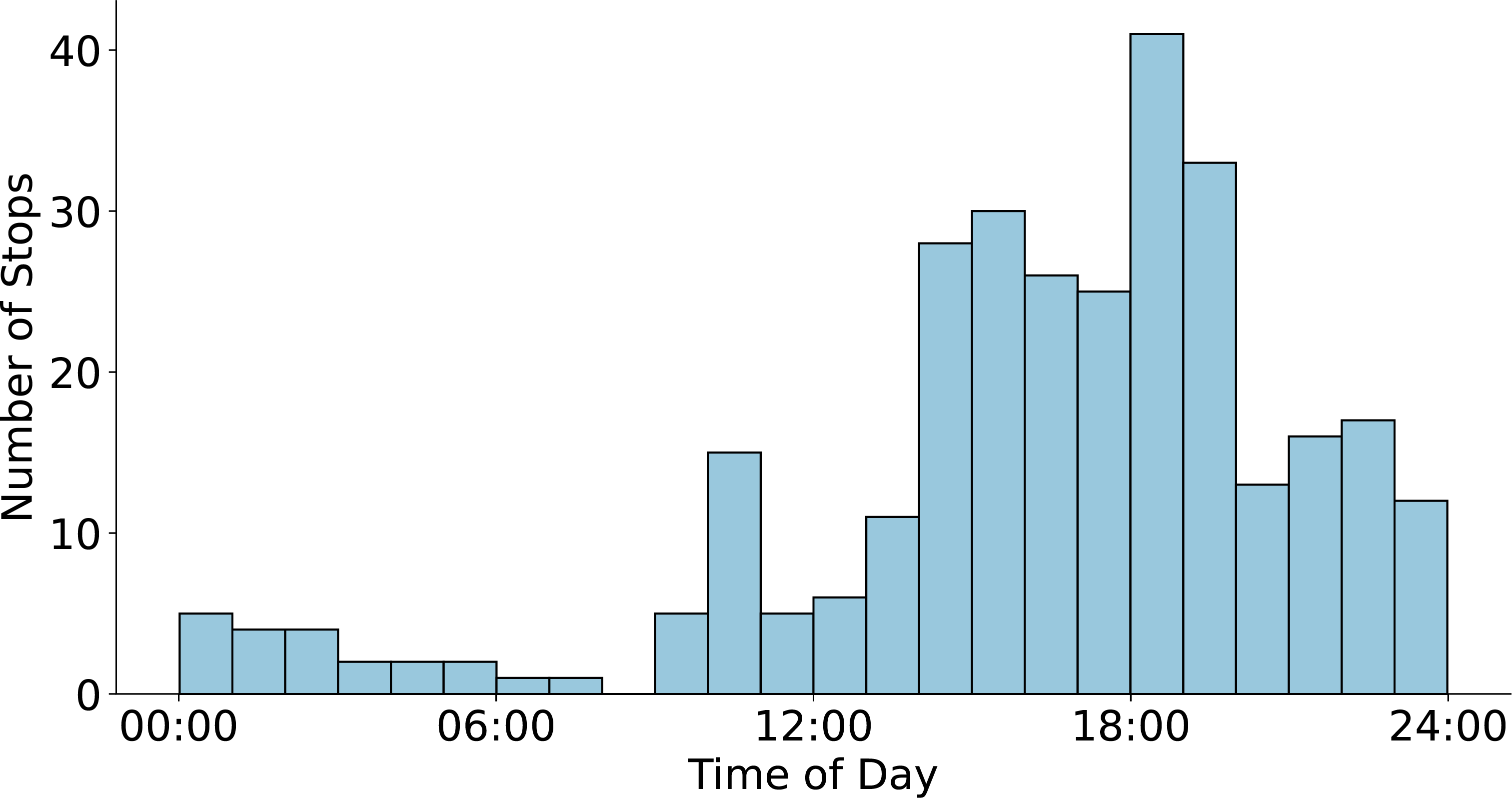}
	\centering
	\caption{Daily stop frequency distribution for a selected individual.}
	\label{fig:stop_history}
\end{figure}

\begin{figure}[ht]
    \includegraphics[width=0.8\textwidth]{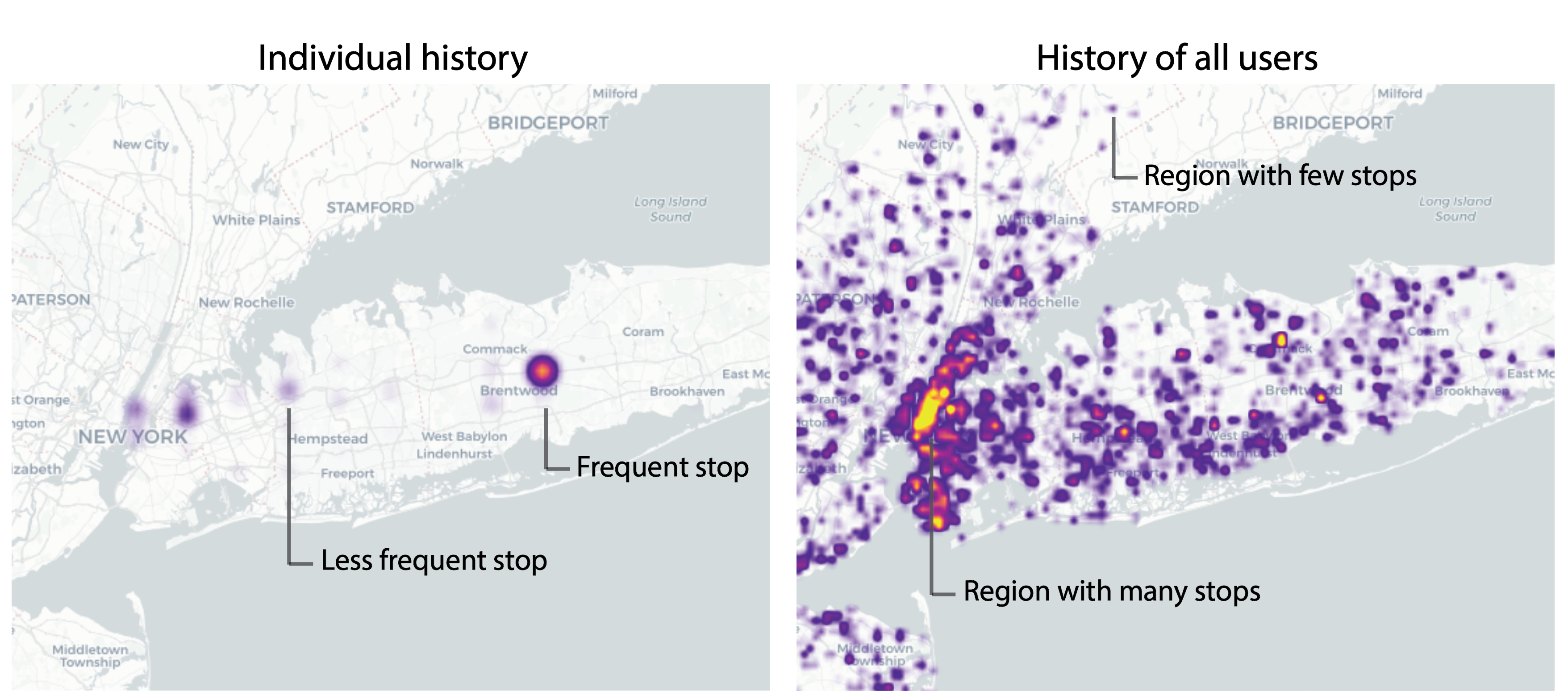}
	\centering
	\caption{Stop density in history of a selected individual (left) and collective (right) history.}
	\label{fig:fig_history_user}
\end{figure}

To characterize our dataset, we provide a set of key measurements along with temporal and geographical visualizations of the data.
Figure~\ref{fig:fig_trend} shows the daily number of unique devices and stops for the entire study period.
A weekly pattern is observable in the data, suggesting the importance of considering the day of the week among the various features.
At the individual level, the daily stop frequency distribution throughout the day is displayed in Figure~\ref{fig:stop_history}: the hours with more stops are the central hours of the day, possibly depicting a work routine habit of the individual.
Community-level data shows broader trends, with a concentration of stops in urban areas, as shown in Figure~\ref{fig:fig_history_user}. Furthermore, the right figure reveals that certain locations exhibit a notably higher concentration of stops, likely due to the presence of points of interest in these areas. The contrast between individual and collective patterns emphasizes the possible benefit of considering both personal routines and collective social behaviors.

\subsection{Processing pipeline}
\begin{figure}[ht]
    \includegraphics[width=\textwidth]{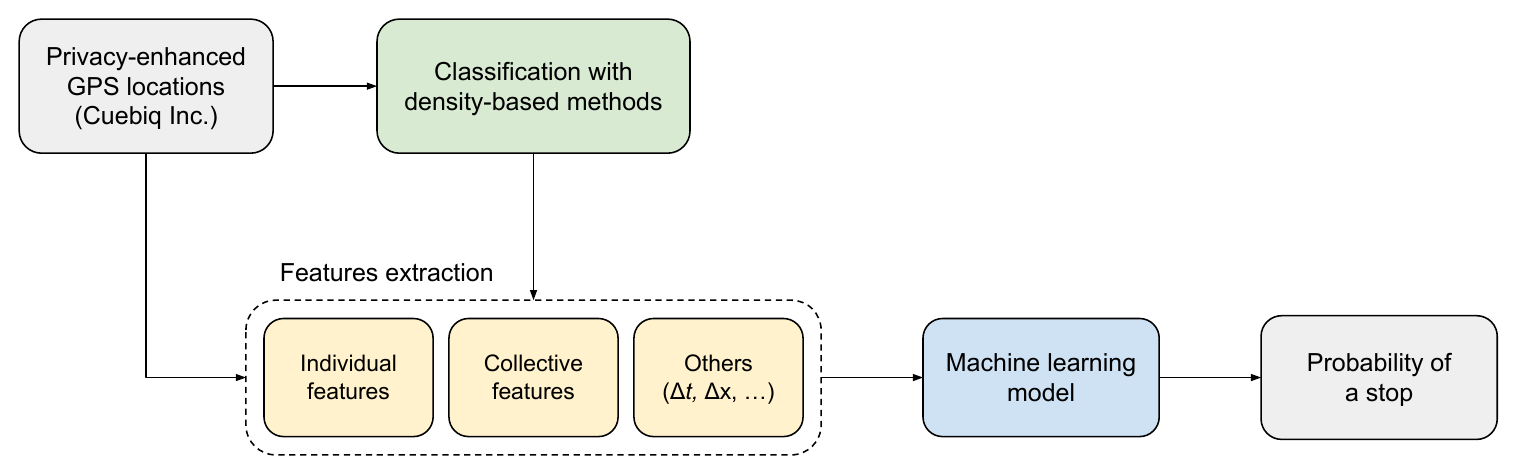}
	\centering
	\caption{Data processing pipeline. The different steps of our method are visualized, including the creation of a labeled dataset and the extraction of relevant features.}
	\label{fig:pipeline}
\end{figure}

First, a density-based algorithm was applied to the Cuebiq data to classify the stopping points according to the presence of clustered GPS datapoints. This creates a labeled dataset that can be used both to extract individual and collective features and to train the machine learning models. Therefore, features about past individual and collective behaviors are extracted, together with some information about the temporal and spatial distance between consecutive points. Then, the labeled features are used in a machine learning model that is trained to determine the probability that the entry point is a stop. The processing pipeline is schematically represented in Figure~\ref{fig:pipeline}.

\begin{figure}[t!]
    \centering
        \includegraphics[width=0.65\textwidth]{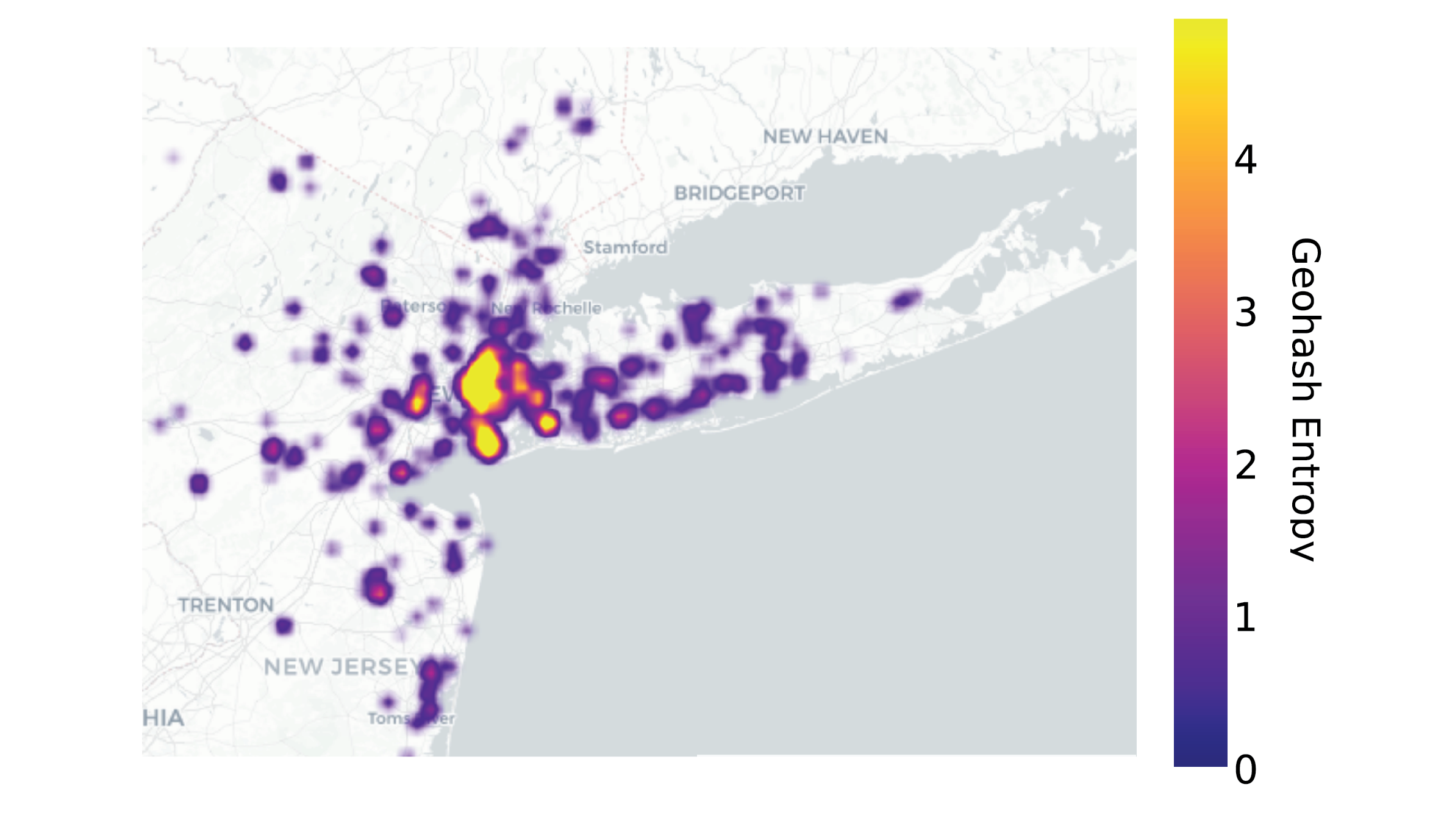}
        	\centering
                \vspace{-5mm}
        	\caption{Geohash entropy: more densely populated areas are associated with higher entropy, corresponding to higher uncertainty in the prediction of mobility patterns.}
        	\label{fig: Entropy_plots}
\end{figure}

\paragraph{Feature engineering}
To detect stop locations at high and low-density points, we integrate several features summarized in Table~\ref{tab:feature}. 
First, we consider individual patterns by computing the number of stops made by the corresponding device at each point in the associated geohash level 8 within the past hour, day, and week.
Then, as suggested by~\citep{Bontorin2024a} equivalent metrics are calculated at the collective level, analyzing the collective behavior within the same geohash area over three different time frames: the previous hour, day, and week.
Along with the temporal features mentioned above, we include recurring historical features considering the number of stops per day of the week and four-hour time blocks (0-3, 4-7, 8-11, 12-15, 16-19, 20-23) in each geohash level 8. This is done both at the individual and community level.
On top of that, we associate each geohash with a measure of entropy to assess the predictability of a geohash in terms of the behavior of individuals in that location. The entropy for each geohash is defined as:
\begin{equation}
    \label{eq:entropy}
    S_{j} = -\sum_{i=1}^n p_{ij} \log \left(p_{ij}\right)\,,
\end{equation}
where $p_{ij}$ is the probability of a stop in the geohash $j$ for the device $i$, computed as the ratio between stops and total passes over the entire observation period. The total number of devices passing in geohash $j$ is $n$.
\begin{table}[b!]
	\centering
 \begin{scriptsize}
	\begin{tabular}{ll}
		\toprule
		\textbf{Feature}     & \textbf{Description}   \\
		\midrule
            Individual routine & \\ 
		\hspace{4mm} Previous hour individual history &  Individual's stop count in a specific geohash during the previous hour \\
		\hspace{4mm} Previous day individual history     & Individual's stop count in a specific geohash during the previous day  \\
		\hspace{4mm} Previous week individual history     & Individual's stop count in a specific geohash during the previous week  \\
		\hspace{4mm} Geohash8 individual history by hour of the day   & Individual's historical stop count in a specific geohash during the same four-hour time window \\
		\hspace{4mm} Geohash8 individual history by day of the week    & Individual's historical stop count in a specific geohash during the same day of the week  \\\midrule
  
            Collective behaviours & \\
		\hspace{4mm} Previous hour collective history & Total stops within a specific geohash during the previous hour \\
		\hspace{4mm} Previous day collective history     & Total stops within a specific geohash during the previous day  \\
		\hspace{4mm} Previous week collective history     & Total stops within a specific geohash during the previous week  \\
		\hspace{4mm} Geohash8 collective history by hour of the day   & Total historical stops in a specific geohash during the same four-hour time window  \\
		\hspace{4mm} Geohash8 collective history by day of the week     & Total historical stops in a specific geohash during the same day of the week  \\
		\hspace{4mm} Geohash8 entropy     & Entropy of the distribution of distinct devices per geohash (Eq.~\ref{eq:entropy})\\\midrule

            Others & \\
            \hspace{4mm} Time interval & Temporal difference between the timestamps of the previous and the following locations for each device \\
		\hspace{4mm} Spatial interval & Spatial distance between the previous and following location for each device \\
	    \hspace{4mm} Cuebiq's point type     & Categorical feature distinguishing between \textit{whitelisted}, \textit{personal\_area} and \textit{other} points \\
	    \hspace{4mm} Accuracy     &  Precision of the device signal as the radius of the circle centered in the coordinate pair given by the ping \\
            \hspace{4mm} Next point accuracy     &  Precision of the device's next signal  \\
            \hspace{4mm} Previous point accuracy     &  Precision of the device's previous signal  \\

		\bottomrule
	\end{tabular}
	\label{tab:feature}
 
 \end{scriptsize}
 \vspace{4mm}
 \caption{Features summary table.}
\vspace{-3.5mm}
\end{table}
Figure~\ref{fig: Entropy_plots} shows the geohash entropy for the considered area. The visualization reveals that geohashes corresponding to more densely populated areas exhibit higher entropy, indicating less predictable movement patterns.
This suggests a relationship between population density and the complexity of movement patterns within each geohash. Highly populated areas likely experience more diverse and frequent movements, leading to increased entropy.
Furthermore, as stop points require stationarity, we incorporate in the features two key quantities related to time and space:
\begin{enumerate}
    \item the temporal distance between the timestamp of the previous and the following location of the same device;
    \item the spatial distance between the previous and the following location of the same device.
\end{enumerate}
The last one is computed with the haversine formula, which determines the great-circle distance between two points on a sphere given their longitudes and latitudes. We also include a categorical feature provided by Cuebiq that differentiates the points in the following classes:
\begin{itemize}
    \item \textit{whitelisted}: any point that falls within a ``whitelist'' point of interest (POI) included in Cuebiq's POI whitelist;
    \item \textit{personal\_area}: top recurrent area for that device;
    \item \textit{other}: any point that does not meet the criteria of the other classifications.
\end{itemize}
Finally, we consider the accuracy of the device signal as a feature, representing the radius of a circle centered in the coordinate pair given by the ping. We consider not only the accuracy of a point but also the accuracy of the previous and following point in the trajectory of the device.

All the numerical features are normalized with a standard scaler such that each has zero mean and unit variance, while one-hot encoding is applied to the categorical ones.

\paragraph{Model evaluation}

We design the model evaluation framework encompassing a training, test, and validation procedure. For this scope, we divide the dataset into training (60\%), validation (20\%), and test (20\%) sets using a temporal ordering to prevent data leakage. The split is based on stop points to ensure each set accurately represents the overall distribution of stops. In addition, this avoids having some stops split across different sets, thus causing data leakage. The resulting process follows these steps:
\begin{enumerate}
    \item it determines a reference date that splits the stop points so that 60\% of them fall before it (or 20\% for validation);
    \item it assigns non-stop points to each set based on their timestamp;
    \item to maintain data integrity, it ensures all points from a single stop event are in the same set, eventually moving the problematic ones (e.g. from the validation set to the training set).    
\end{enumerate}

This approach preserves the temporal structure of the data while maintaining representative proportions of stop and non-stop points across all sets.

\section{Results}
We conduct a preliminary analysis to study the correlation between the various features we consider (subsection~\ref{par:corr}).
Then, we tackle the stop detection problem in case of missing data by training three different classification algorithms: LightGBM Classification, Random Forest, and 3-layers Feed-Forward Neural Network (FFNN).
We employ three different algorithms to develop a more robust and versatile solution that can adapt to various data scenarios, potentially improving accuracy across diverse user groups.
Furthermore, our dataset is imbalanced, containing the majority of data points representing trajectory points rather than stops. This type of data leads us to evaluate the result of the model with measures such as Area Under the ROC Curve (AUC), recall, and precision, rather than traditional metrics such as accuracy. The use of these metrics allows us to correctly measure the performance of our method in the stop-detection task.

In subsection~\ref{par:performance} and Table~\ref{tab:results}, we summarize the values obtained in these metrics by the three models considered. In addition, to better understand the nature of false positives, in subsection~\ref{par:false_pos}, we show an analysis comparing them with true negatives, i.e. locations that both our method and the density-based algorithm classify as ``not stops''.
Finally, in subsection~\ref{par:feature},we show an explainability analysis to have an insight into the features' importance.

\subsection{Between-feature correlation analysis}\label{par:corr}
\begin{figure}[t!]
        \hspace{15mm}
\includegraphics[width=0.7\textwidth]{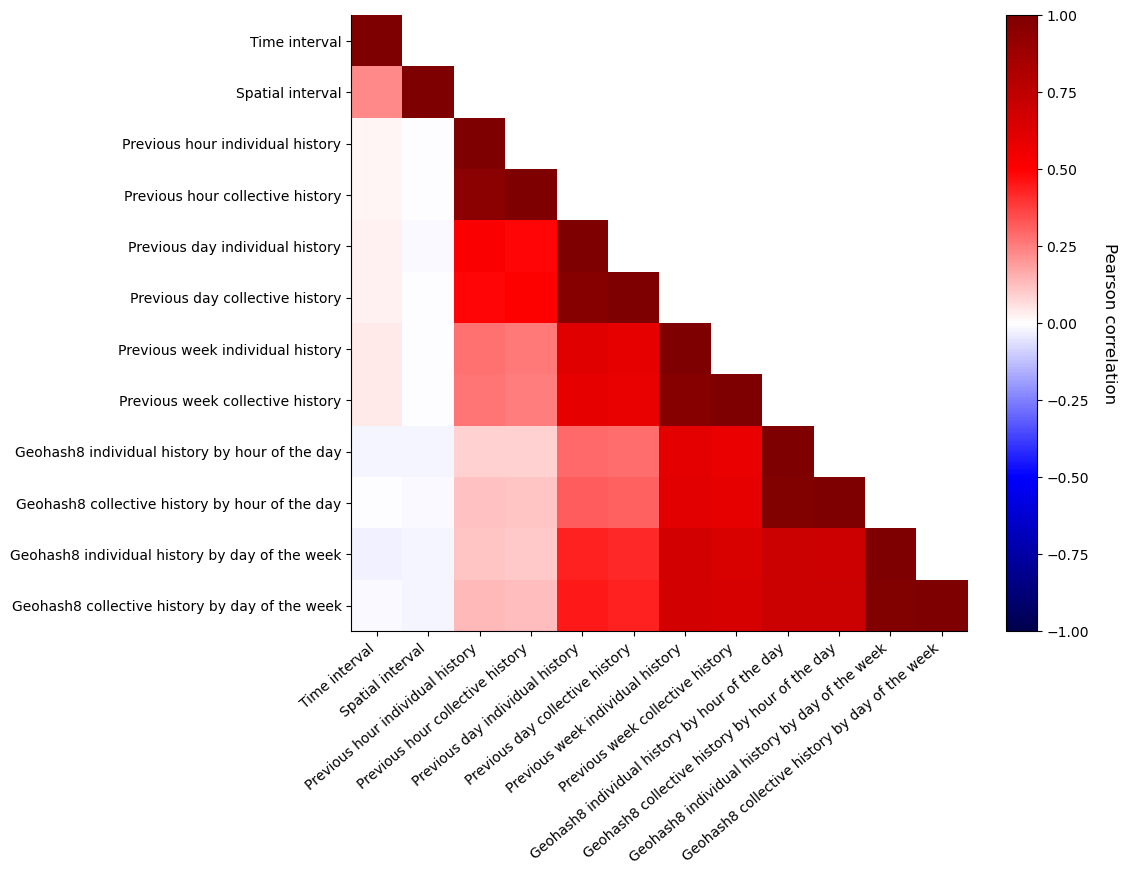}
	\caption{Correlation matrix of the relevant features given by the Pearson coefficient between each pair of variables. The color represents the magnitude of the correlation.}
	\label{fig:correlation_matrix}
\end{figure}

Our feature selection process focuses on identifying variables that indicate a higher probability of a point belonging to a stop. Most of these features incorporate some aspects of spatial and/or temporal proximity to known stops, suggesting a high likelihood of inter-feature correlation. Figure~\ref{fig:correlation_matrix} shows the collinearity between every pair of the most correlated features, all of them computed from the training data and colored according to the magnitude of the Pearson coefficient. As expected, most correlations are positive, and looking at the diagonal we can see groups of very correlated features.

The first pair of variables that correlate are the time and space intervals. This is expected, as for any device moving at comprehensive speeds the longer the time between pings, the higher the space it covers. Both of these features are important as the model takes into account the speed of the device to guess whether it stopped or not. Moreover, the time variable is a useful metric to indicate where there are missing points in a trajectory whether the reason is bad connectivity (for example the device entering a closed space) or the phone having been turned off.

In addition, there is a larger cluster of correlated features which are related to the number of stops near the point in question, whether they belong to the same device or the collective, during certain windows of time. The highest degree of collinearity is observed between pairs of individual-collective stop histories. This phenomenon can be attributed to the relatively small number of unique devices (n = 2,732) included in the feature calculation across the New York-Newark-Jersey City metropolitan area. The low probability of overlapping stops from distinct devices results in the majority of collective feature information encountered by an individual being derived from their own data. Consequently, we opted to neglect the collective features for model training.

\subsection{Performance evaluation}\label{par:performance}

Given the dataset's imbalance, we are more interested in AUC and recall than precision in evaluating the performance. A high recall indicates that our method successfully detects the majority of subsampled stops. On the other hand, the precision is less informative since a small number of false negatives compared to the true negatives can still produce a low precision when compared to the true positives.
In Table~\ref{tab:results}, we reported the results obtained for the three models considered.

The Random Forest demonstrates the highest AUC and recall. However, the performance of all models is notably comparable and validates our approach, providing flexibility in model selection. From this perspective, the most suitable model based on specific use cases or computational requirements can be chosen without compromising performance. 

As expected, all models exhibited excellent AUC but a relatively low precision, indicating that we are identifying more potential stops than the density-based algorithm. This approach ensures we do not miss critical locations, albeit at the cost of some over-identification from points near actual stops. To prove this, we conducted an in-depth analysis of false positive points to better understand their characteristics (details in subsection~\ref{par:false_pos}). 

\begin{table}[H]
	\centering
	\begin{tabular}{l|rrr}
		\toprule
		\textbf{Algorithm}  & \textbf{ROC AUC} & \textbf{Recall} & \textbf{Precision}\\
		\midrule
		LightGBM Classification & 0.980 & 0.760 & 0.027\\
 		Random Forest  & 0.978 & 0.809 & 0.026\\
		1-layer FFNN & 0.972 &  0.757 & 0.023\\
  
		\bottomrule
	\end{tabular}
	
 \vspace{3.5mm}
 \caption{Results summary table. The three models considered achieve very similar performance, with high AUC, recall and low precision, caused by the dataset's imbalance.}
 \label{tab:results}
\vspace{-3.5mm}
\end{table}

\subsection{Comparative analysis of False Positives}\label{par:false_pos}
For each of the 18174 false positives (incorrectly identified stop), we calculate the haversine distance in meters to the nearest actual stop location of the same individual.
This analysis aims to understand the spatial relationship between misclassified points and actual stops, potentially revealing patterns in our algorithm's performance.
We perform the same distance calculation for true negatives (correctly identified non-stops), randomly sampling 20000 points from the total 116596 true negatives for the analysis.
Table~\ref{tab:tn_fp_analysis} presents the quantiles of the distance distribution to the nearest stop for both false positives and true negatives. Our analysis reveals that false positive points tend to be closer to actual stop locations compared to true negative points.
Indeed, most of the locations identified as false positives are recurring points for the device, situated near actual past stop points for the device.
Figure~\ref{fig: False_positives} provides a case study, illustrating false positive points (in red) nearby actual stop locations (in green). This visualisation helps provide context to the spatial relationship between misclassified and real stops.

 \begin{minipage}[b]{0.5\textwidth}
    \centering
        \begin{table}[H]
        	\centering
        	\begin{tabular}{l|r|r}
        		\toprule
        		& \textbf{False Positives [m]}  & \textbf{True Negatives [m]}  \\
        		\midrule
                Min  & 0.00 & 0.00 \\ 
                25\% & 2.11 & 410.36 \\ 
                50\% & 7.23 & 1409.49 \\ 
                75\% & 23.24 & 3923.87 \\ 
                Max  & 73961.78 & 72413.30 \\ 
        		\bottomrule
        	\end{tabular}
        	\label{tab:tn_fp_analysis} 
        \vspace{3.5mm}
        \caption{Minimum distance distribution in meters to the nearest stop at the device level for both false positive resulting points and for the true negative resulting points.}
        \vspace{-3.5mm}
        \end{table}
\end{minipage}
\hspace{0.05\textwidth} 
\begin{minipage}[b]{0.40\textwidth}
    \centering
        \begin{figure}[H]
        \includegraphics[width=0.885\textwidth]{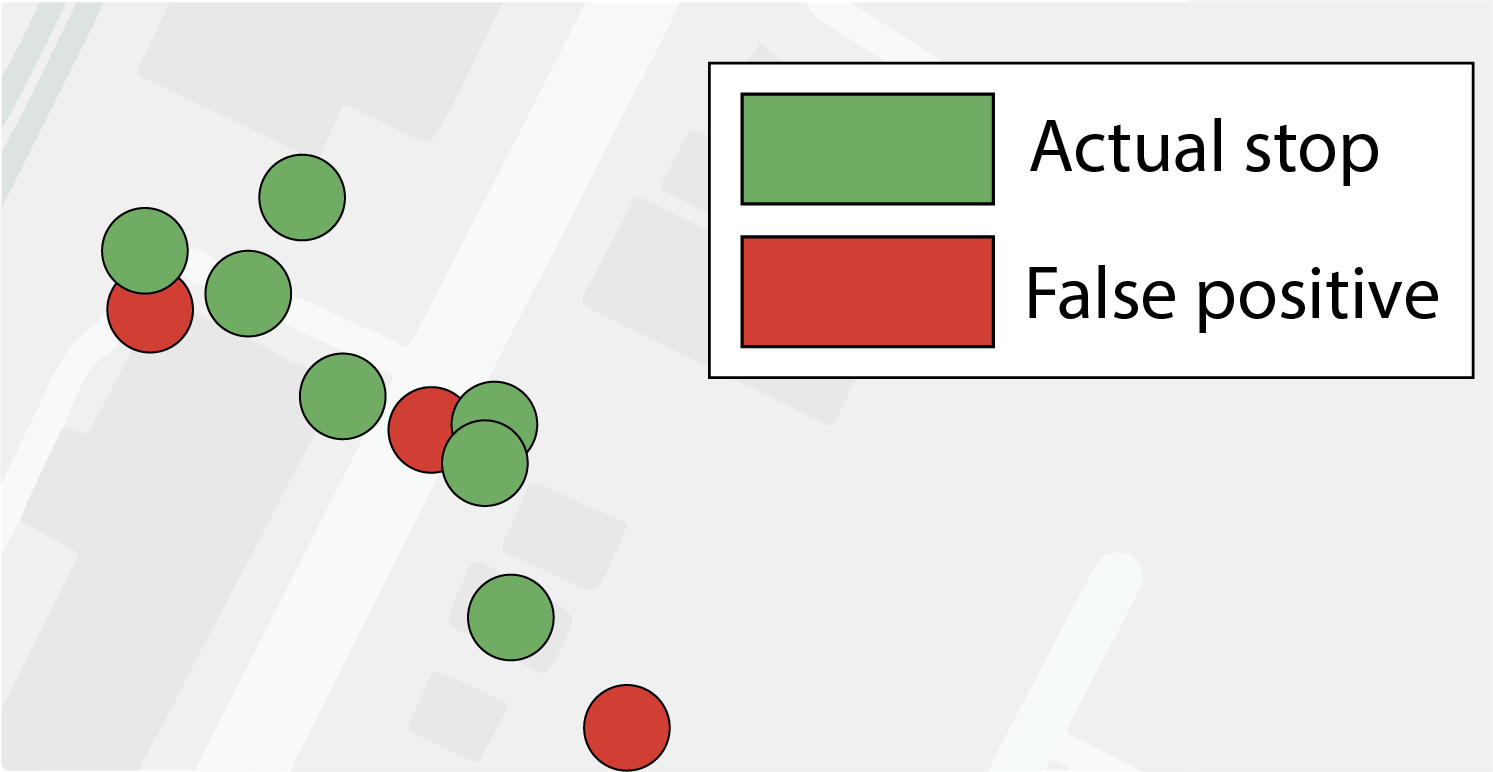}
        	\centering
        	\caption{Case study: false positive points (red) and actual stop points (green).}
        	\label{fig: False_positives}
        \end{figure}
\end{minipage}

\subsection{Feature importance}\label{par:feature}

To assess the effectiveness of our approach, we analyze feature importance, enhancing our ability to interpret and explain the results. This approach allows us to understand better which factors contribute mostly to accurate stop location detection.

Our analysis of the features reveals that the most critical features for detecting stop locations are the time interval and space interval associated with each device position. These features aid in identifying non-stationary points, aligning with the latest algorithms that initially filter out moving points~\citep{Ramaswamy2004}. Another important feature is the number of individual stops made in the previous week, which leverages the routine patterns of individuals. Additionally, location accuracy plays a determining role, as it reflects the reliability of a point's position based on device signal precision. This finding underscores the importance of a point's actual position in determining whether it's a stop location.

Our comprehensive feature analysis reveals that collective measures have minimal impact on our models, primarily due to the limited sample size. The small number of devices in our study prevented us from uncovering meaningful collective trends. This limitation is evident in the feature correlation analysis in subsection~\ref{par:corr}.

\subsection{Limitations}\label{par:limitation}
The ground truth dataset we used in this research is obtained through a density-based clustering algorithm. Therefore, we do not know whether the stops in the ground truth data belong to actual stops or whether the clustering algorithm misclassified them. To enhance the quality of our work, a more reliable dataset with verified stop classifications would be valuable, as it would prevent training the model on potentially erroneous data and would make the results more reliable. One approach to obtain such verified data could involve test users manually recording their actual stops.

In addition, our dataset is inherently imbalanced, with most data points representing movement rather than stops. This imbalance is characteristic of human mobility patterns, where people typically traverse more locations than they stop at, putting constraints on the types of performance metrics that can be used.

Furthermore, in the modeling part, we could not fully exploit the original data due to constraints on time and computational resources. Consequently, we downsampled the dataset to cover two months instead of four months and 121 devices. Performing the analysis again with the larger dataset could improve the performance, as more data could be used for training the model.
Also, the limited number of devices in our current study has constrained our ability to fully explore the impact of collective patterns. Future research with a larger device sample has the potential to enhance our models' precision performance by leveraging these broader behavioral trends.

\section{Conclusions}
In this research, we addressed the problem of identifying stop locations even when data points are missing. Indeed, while the use of GPS data opens up new possibilities for analyzing large-scale movement data, it also poses the problem of managing gaps in the data collection. To tackle the problem, we considered three models, training them with features encoding individual routines and collective mobility behavior, plus temporal and spatial information between consecutive moves. 

All our models demonstrate comparable performances, successfully identifying a high number of correct stop locations. Our comparative analysis of false positives revealed that points incorrectly labeled as stops by our algorithm are, in fact, recurring locations for the device, often situated near actual past stop points. 
This finding validates the effectiveness of our chosen features in identifying potential stop locations.

Due to computational constraints, we were limited in the number of devices we could analyze, which prevented us from understanding the potential impact of collective patterns. Future work on the same dataset including a wider range of devices for the training part of the models has the potential to further improve our results.

To further enhance our model, some external factors such as weather conditions and public holidays can complement the feature set. These elements could specifically influence some individual routines, affecting personal decisions, and shaping collective mobility patterns routines related to social events and mobility transportation. Moving forward, a hybrid approach that combines the strengths of multiple models might be designed, potentially improving overall accuracy and reliability in stop detection.

\section{Acknowledgements}
This work is the output of the Complexity72h workshop, held at the Universidad Carlos III de Madrid in Leganés, Spain, 24-28 June 2024. https://www.complexity72h.com

The authors would like to thank Cuebiq for kindly providing us with the mobility dataset for this research through their Data for Good program.

\bibliographystyle{unsrtnat}
\bibliography{enhancing_stop_detection}  

\begin{thebibliography}{19}
\providecommand{\natexlab}[1]{#1}
\providecommand{\url}[1]{\texttt{#1}}
\expandafter\ifx\csname urlstyle\endcsname\relax
  \providecommand{\doi}[1]{doi: #1}\else
  \providecommand{\doi}{doi: \begingroup \urlstyle{rm}\Url}\fi

\bibitem[De~Nadai et~al.(2016)De~Nadai, Staiano, Larcher, Sebe, Quercia, and
  Lepri]{DeNadai2016}
Marco De~Nadai, Jacopo Staiano, Roberto Larcher, Nicu Sebe, Daniele Quercia,
  and Bruno Lepri.
\newblock The death and life of great italian cities: A mobile phone data
  perspective.
\newblock In \emph{Proceedings of the 25th International Conference on World
  Wide Web}, WWW '16, page 413–423, Republic and Canton of Geneva, CHE, 2016.
  International World Wide Web Conferences Steering Committee.
\newblock ISBN 9781450341431.
\newblock \doi{10.1145/2872427.2883084}.
\newblock URL \url{https://doi.org/10.1145/2872427.2883084}.

\bibitem[Ferretti et~al.(2018)Ferretti, Barlacchi, Pappalardo, Lucchini, and
  Lepri]{Ferretti2018}
Michele Ferretti, Gianni Barlacchi, Luca Pappalardo, Lorenzo Lucchini, and
  Bruno Lepri.
\newblock Weak nodes detection in urban transport systems: Planning for
  resilience in singapore.
\newblock In \emph{2018 IEEE 5th International Conference on Data Science and
  Advanced Analytics (DSAA)}, pages 472--480, 2018.
\newblock \doi{10.1109/DSAA.2018.00061}.

\bibitem[Cottrill and Derrible(2015)]{Caitlin2015}
Caitlin~D. Cottrill and Sybil Derrible.
\newblock Leveraging big data for the development of transport sustainability
  indicators.
\newblock \emph{Journal of Urban Technology}, 22\penalty0 (1):\penalty0 45--64,
  2015.
\newblock \doi{10.1080/10630732.2014.942094}.
\newblock URL \url{https://doi.org/10.1080/10630732.2014.942094}.

\bibitem[Kraemer et~al.(2020)Kraemer, Yang, Gutierrez, Wu, Klein, Pigott,
  Group†, du~Plessis, Faria, Li, Hanage, Brownstein, Layan, Vespignani, Tian,
  Dye, Pybus, and Scarpino]{Moritz2020}
Moritz U.~G. Kraemer, Chia-Hung Yang, Bernardo Gutierrez, Chieh-Hsi Wu, Brennan
  Klein, David~M. Pigott, Open COVID-19 Data~Working Group†, Louis
  du~Plessis, Nuno~R. Faria, Ruoran Li, William~P. Hanage, John~S. Brownstein,
  Maylis Layan, Alessandro Vespignani, Huaiyu Tian, Christopher Dye, Oliver~G.
  Pybus, and Samuel~V. Scarpino.
\newblock The effect of human mobility and control measures on the covid-19
  epidemic in china.
\newblock \emph{Science}, 368\penalty0 (6490):\penalty0 493--497, 2020.
\newblock \doi{10.1126/science.abb4218}.
\newblock URL \url{https://www.science.org/doi/abs/10.1126/science.abb4218}.

\bibitem[Oliver et~al.(2020)Oliver, Lepri, Sterly, Lambiotte, Deletaille,
  Nadai, Letouzé, Salah, Benjamins, Cattuto, Colizza, de~Cordes, Fraiberger,
  Koebe, Lehmann, Murillo, Pentland, Pham, Pivetta, Saramäki, Scarpino,
  Tizzoni, Verhulst, and Vinck]{Oliver2020}
Nuria Oliver, Bruno Lepri, Harald Sterly, Renaud Lambiotte, Sébastien
  Deletaille, Marco~De Nadai, Emmanuel Letouzé, Albert~Ali Salah, Richard
  Benjamins, Ciro Cattuto, Vittoria Colizza, Nicolas de~Cordes, Samuel~P.
  Fraiberger, Till Koebe, Sune Lehmann, Juan Murillo, Alex Pentland, Phuong~N
  Pham, Frédéric Pivetta, Jari Saramäki, Samuel~V. Scarpino, Michele
  Tizzoni, Stefaan Verhulst, and Patrick Vinck.
\newblock Mobile phone data for informing public health actions across the
  covid-19 pandemic life cycle.
\newblock \emph{Science Advances}, 6\penalty0 (23):\penalty0 eabc0764, 2020.
\newblock \doi{10.1126/sciadv.abc0764}.
\newblock URL \url{https://www.science.org/doi/abs/10.1126/sciadv.abc0764}.

\bibitem[Aguilar-Sánchez et~al.()Aguilar-Sánchez, Bassolas, Ghoshal, Hazarie,
  Kirkley, Mazzoli, Meloni, Mimar, Nicosia, Ramasco, and
  Sadilek]{Aguilar-Sanchez2022}
Javier Aguilar-Sánchez, Aleix Bassolas, Gourab Ghoshal, Surendra~A. Hazarie,
  Alec Kirkley, Mattia Mazzoli, Sandro Meloni, Sayat Mimar, Vizcenzo Nicosia,
  José~J. Ramasco, and Adam Sadilek.
\newblock Impact of urban structure on infectious disease spreading.
\newblock ISSN 2045-2322.
\newblock \doi{10.13039/100000001}.
\newblock URL \url{https://digital.csic.es/handle/10261/269380}.

\bibitem[Klein et~al.(2023)Klein, Zenteno, Joseph, Zahedi, Hu, Copenhaver,
  Kraemer, Chinazzi, Klompas, Vespignani, et~al.]{klein2023forecasting}
Brennan Klein, Ana~C Zenteno, Daisha Joseph, Mohammadmehdi Zahedi, Michael Hu,
  Martin~S Copenhaver, Moritz~UG Kraemer, Matteo Chinazzi, Michael Klompas,
  Alessandro Vespignani, et~al.
\newblock Forecasting hospital-level covid-19 admissions using real-time
  mobility data.
\newblock \emph{Communications Medicine}, 3\penalty0 (1):\penalty0 25, 2023.

\bibitem[Pappalardo et~al.(2015)Pappalardo, Pedreschi, Smoreda, and
  Giannotti]{Pappalardo2015}
Luca Pappalardo, Dino Pedreschi, Zbigniew Smoreda, and Fosca Giannotti.
\newblock Using big data to study the link between human mobility and
  socio-economic development.
\newblock In \emph{2015 IEEE International Conference on Big Data (Big Data)},
  pages 871--878, 2015.
\newblock \doi{10.1109/BigData.2015.7363835}.

\bibitem[Centellegher et~al.(2024)Centellegher, De~Nadai, Tonin, Lepri, and
  Lucchini]{Centellegher2024}
Simone Centellegher, Marco De~Nadai, Marco Tonin, Bruno Lepri, and Lorenzo
  Lucchini.
\newblock The long-term and disparate impact of job loss on individual mobility
  behaviour, 2024.
\newblock URL \url{http://arxiv.org/abs/2403.10276}.

\bibitem[Hariharan and Toyama(2004)]{Ramaswamy2004}
Ramaswamy Hariharan and Kentaro Toyama.
\newblock Project lachesis: Parsing and modeling location histories.
\newblock In Max~J. Egenhofer, Christian Freksa, and Harvey~J. Miller, editors,
  \emph{Geographic Information Science}, pages 106--124, Berlin, Heidelberg,
  2004. Springer Berlin Heidelberg.
\newblock ISBN 978-3-540-30231-5.

\bibitem[Aslak and Alessandretti(2020)]{Aslak2020}
Ulf Aslak and Laura Alessandretti.
\newblock Infostop: {{Scalable}} stop-location detection in multi-user mobility
  data, 03 2020.
\newblock URL \url{http://arxiv.org/abs/2003.14370}.

\bibitem[Ester et~al.(1996)Ester, Kriegel, Sander, Xu,
  et~al.]{ester1996density}
Martin Ester, Hans-Peter Kriegel, J{\"o}rg Sander, Xiaowei Xu, et~al.
\newblock A density-based algorithm for discovering clusters in large spatial
  databases with noise.
\newblock In \emph{kdd}, volume~96, pages 226--231, 1996.

\bibitem[Ankerst et~al.(1999)Ankerst, Breunig, Kriegel, and
  Sander]{ankerst1999optics}
Mihael Ankerst, Markus~M Breunig, Hans-Peter Kriegel, and J{\"o}rg Sander.
\newblock Optics: Ordering points to identify the clustering structure.
\newblock \emph{ACM Sigmod record}, 28\penalty0 (2):\penalty0 49--60, 1999.

\bibitem[Palma et~al.(2008)Palma, Bogorny, Kuijpers, and
  Alvares]{palma2008clustering}
Andrey~Tietbohl Palma, Vania Bogorny, Bart Kuijpers, and Luis~Otavio Alvares.
\newblock A clustering-based approach for discovering interesting places in
  trajectories.
\newblock In \emph{Proceedings of the 2008 ACM symposium on Applied computing},
  pages 863--868, 2008.

\bibitem[Buchin et~al.(2011)Buchin, Driemel, Van~Kreveld, and
  Sacrist{\'a}n]{buchin2011segmenting}
Maike Buchin, Anne Driemel, Marc~J Van~Kreveld, and Vera Sacrist{\'a}n.
\newblock Segmenting trajectories: A framework and algorithms using
  spatiotemporal criteria.
\newblock \emph{Journal of Spatial Information Science}, 3:\penalty0 33--63,
  2011.

\bibitem[Soares~J{\'u}nior et~al.(2015)Soares~J{\'u}nior, Moreno, Times,
  Matwin, and Cabral]{soares2015grasp}
Am{\'\i}lcar Soares~J{\'u}nior, Bruno~Neiva Moreno, Val{\'e}ria~Ces{\'a}rio
  Times, Stan Matwin, and Luc{\'\i}dio dos Anjos~Formiga Cabral.
\newblock Grasp-uts: an algorithm for unsupervised trajectory segmentation.
\newblock \emph{International Journal of Geographical Information Science},
  29\penalty0 (1):\penalty0 46--68, 2015.

\bibitem[Bontorin et~al.()Bontorin, Centellegher, Gallotti, Pappalardo, Lepri,
  and Luca]{Bontorin2024a}
Sebastiano Bontorin, Simone Centellegher, Riccardo Gallotti, Luca Pappalardo,
  Bruno Lepri, and Massimiliano Luca.
\newblock Mixing {{Individual}} and {{Collective Behaviours}} to {{Predict
  Out-of-Routine Mobility}}.
\newblock URL \url{https://arxiv.org/abs/2404.02740v1}.

\bibitem[Klein et~al.()Klein, Zenteno, Joseph, Zahedi, Hu, Copenhaver, Kraemer,
  Chinazzi, Klompas, Vespignani, Scarpino, and Salmasian]{Klein2023}
Brennan Klein, Ana~C. Zenteno, Daisha Joseph, Mohammadmehdi Zahedi, Michael Hu,
  Martin~S. Copenhaver, Moritz U.~G. Kraemer, Matteo Chinazzi, Michael Klompas,
  Alessandro Vespignani, Samuel~V. Scarpino, and Hojjat Salmasian.
\newblock Forecasting hospital-level {{COVID-19}} admissions using real-time
  mobility data.
\newblock 3\penalty0 (1):\penalty0 25.
\newblock ISSN 2730-664X.
\newblock \doi{10.1038/s43856-023-00253-5}.

\bibitem[Niemeyer()]{geohash}
G.~Geohash. Niemeyer.
\newblock Geohash.
\newblock \url{https://en.wikipedia.org/wiki/Geohash}.
\newblock Accessed: 2008.

\end{thebibliography}






\end{document}